\documentclass[a4paper,10pt]{article}
\usepackage[utf8]{inputenc}
\usepackage{amsmath}
\usepackage{cite}
\usepackage{float}
\usepackage{amsfonts}
\usepackage{graphicx}
\usepackage{pbox}

\title{Capacity Studies for a Differential Growing Neural Gas}
\author{P. Levi, P. Gelhausen, G. Peters}

\begin{document}

\maketitle

\begin{abstract}

In 2019 Kerdels and Peters proposed a grid cell model (GCM) based on a Differential Growing Neural Gas (DGNG) network architecture as a computationally efficient way to model an Autoassociative Memory Cell (AMC) \cite{Kerdels_Peters_2019}. An important feature of the DGNG architecture with respect to possible applications in the field of computational neuroscience is its \textit{capacity} refering to its capability to process and uniquely distinguish input signals and therefore obtain a valid representation of the input space. This study evaluates the capacity of a two layered DGNG grid cell model on the Fashion-MNIST dataset. The focus on the study lies on the variation of layer sizes to improve the understanding of capacity properties in relation to network parameters as well as its scaling properties. Additionally, parameter discussions and a plausability check with a pixel/segment variation method are provided. It is concluded, that the DGNG model is able to obtain a meaningful and plausible representation of the input space and to cope with the complexity of the Fashion-MNIST dataset even at moderate layer sizes.

\end{abstract}

\section{Introduction}

The discovery of entorhinal \textit{grid cells} in the brains of several mammelians (including humans) by \cite{Moser_2005} in 2005 sparked a wave of new research and modeling approaches in the field of neuroscience. It was found that their individual neuronal activity correlates strongly with periodic patterns in the environment of a traversing animal \cite{Fyhn_2004}. Therefore in many models grid cells are hypothesized to be part of a specialized system for navigation and orientation \cite{Rowland_2016}, processing and storing information about location and movement vectors. Yet their exact purpose and function is still not fully understood at this point.

From a more general point of view, it was shown that the grid cell activity pattern can also be interpreted as an instance of a more general information processing scheme \cite{Kerdels_Peters_2015, Kerdels_Peters_2016}. The authors propose a grid cell model (GCM) based on a Recursive Growing Neural Gas (RGNG), which is then modified and extended to model an Autoassociative Memory Cell (AMC). In \cite{Kerdels_Peters_2018}, the basic capabilities of this AMC were demonstrated in a preliminary simulation using the MNIST dataset. Kerdels and Peters in \cite{Kerdels_Peters_2019} further introduced a model based on a Differential Growing Neural Gas (DGNG) as an approximation to the RGNG model. The DGNG approximation showed more efficient computational properties while largely maintaining the desired performance.
These encouraging results revealed the need for further research regarding specific aspects of the model, especially its \textit{capacity}. In this context, capacity is understood as the model's ability of mapping input signals injectivly on a set of output. More specifically, this refers to the ability of the model to distinguish a variety of input signals and therefore obtain a valid representation of the input space. This input signals can for example correspond to sensory input from images. The capacity is an important attribute in the context of computational neuroscience. Every candidate model should be able to enable processing (and abstraction) of information from sensory input to a certain degree. On the other side, the model should not ignore or discard information relevant to the use case at hand. The purpose of this study is to probe the capacity of the proposed DGNG model \cite{Kerdels_Peters_2019}, as well as corresponding scaling properties. To this end, an extended capacity study on the more complex Fashion-MNIST dataset is conducted. This work discusses dataset related capacity properties as well as model related ones (including hyperparameters and architecture). As a result, a better understanding of the chances and limitations of the DGNG based model approach is obtained, as well as incentives for future research.

The structure of the paper is as follows: In Section \ref{section_architecture}, the architecture of the DGNG model is briefly discussed. Section \ref{section_methodology} defines similarity metrics and the method for capacity quantification employed in this paper. Section \ref{section_Fashion_MNIST} proceeds to lay out the properties of the Fashion-MNIST dataset for the purpose of training and testing. After a short parameter study and plausability check in Section \ref{section_parameter_analysis} the results of the capacity analysis are finally presented in Section \ref{section_analysis}. The paper closes on a conclusion and outlook in Section \ref{section_conclusion}.

\section{DGNG model architecture}\label{section_architecture}

The architecture of the DGNG model is based on \cite{Kerdels_Peters_2019}. A compact overview of the used model structure is given as well as the conceptional development steps. For a more detailed view, the reader is refered to \cite{Kerdels_Peters_2019}, for reasons of consistency the notation from Kerdels and Peters is adopted throughout this study.

\subsection{The RGNG model}\label{section_RGNG}

The RGNG marks the conceptional starting point of the model development. The RGNG is a unsupervised learning algorithm, which segments the provided input space into \textit{prototypes}. As stated in \cite{Kerdels_Peters_2019}, the RGNG can be ``interpreted as describing the behavior of a group of neurons that receives signals from a common input space''. The formal structure of the RGNG is extensivly described in \cite{Kerdels_Peters_2016, Kerdels_Peters_2019} and is repeated here only in the most important steps:

The RGNG $g$ is defined by the tuple

\begin{align}
 g := (U, C, \theta) \in G,
\end{align}

with a set of units $U$, the edges $C$ and a set of parameters $\theta$. $G$ denotes the set of all RGNGs. The units $u \in U$ of the RGNG are corresponding to the prototypes $w$:

\begin{align}
 u := (w, e, s) \in U,\quad w\in W:= \mathbb{R}^n\cup G, \quad e \in \mathbb{R},
\end{align}

with the accumulated error $e$ of the prototype. The accumulated error describes the deviation of an input vector from the prototype at hand, usually via squared distance. As noted in \cite{Kerdels_Peters_2019}, the prototype $w$ can be either a $n$-dimensional input vector or another RGNG $s$. In the case at hand, the network has a depth of one, resulting in a two layered structure: Each of the units in the first layer $L_1$ corresponds to a sub-RGNG $s$ with units in the bottom layer $L_2$.

The edges of $g$ are also defined as a tupel:

\begin{align}
 C := (V, t) \in C,\quad V\subseteq U \land |V| = 2, \quad t \in \mathbb{N},
\end{align}

with the units $v \in V$ connected by the edge and its age $t$. The age of the edges serves as a treshold parameter for removing obsolete units and edges in the learning process. The dynamics of the network are described in four additional functions:

\begin{itemize}
 \item The \textit{distance function} $D(x,y)$ describes the distance between two parameters $x$ and $y$, which can be both a vector or a (sub)RGNG.
 \item The \textit{interpolation function} $I(x,y)$, which creates a new vector or (sub)RGNG by interpolating between two vectors or RGNGs.
 \item The \textit{adaption function} $A(x,\xi,r)$, which adapts an vector or RGNG $x$ towards the input vector $\xi$ by a given fraction $r$.
 \item The \textit{input function} $F(g,\xi)$, which provides the distance between the input vector $\xi$ and the best matching unit (BMU) of the RGNG $g$. It also provides the adapted RGNG $g$ after the process.  
\end{itemize}

As an interpretation in terms of a neuronal model, the units in the first layer can be seen as the neurons, while the sub-RGNGS in the bottom layers correspond to their dendritic trees, in which the prototypical input patterns are stored. Because the neurons are in competition during the learning process, they learn different representations. Although the emerging prototypes are pairwise distinct from each other, in sum they still form a coarse representation of the input space. In an abstract sense, each neuron offers an own ``perspective'' of the input space, and taking into account all single perspectives, the whole picture can be reconstructed.

\subsection{The RGNG approximation}\label{section_DGNG}

The RGNG model has been successfully used to describe possible behavior of entorhinal (grid) cells (e.g. \cite{Kerdels_Peters_2016, Kerdels_Peters_2019_2}). As a central part the algorithm requires the distance calculation of the input vector to \textit{each} prototypical input pattern of \textit{each} neuron. While feasible for small neuron groups, this can easily exceed the capacity of modern systems when scaling up to larger setups. Therefore Kerdels and Peters introduced an approximation to the RGNG called \textit{Differential Growing Neural Gas} (DGNG). Again, the description in full detail is given in \cite{Kerdels_Peters_2019} and is only repeated in the key points important for this study.

\begin{figure}[H]
\centering
\includegraphics[scale=0.2]{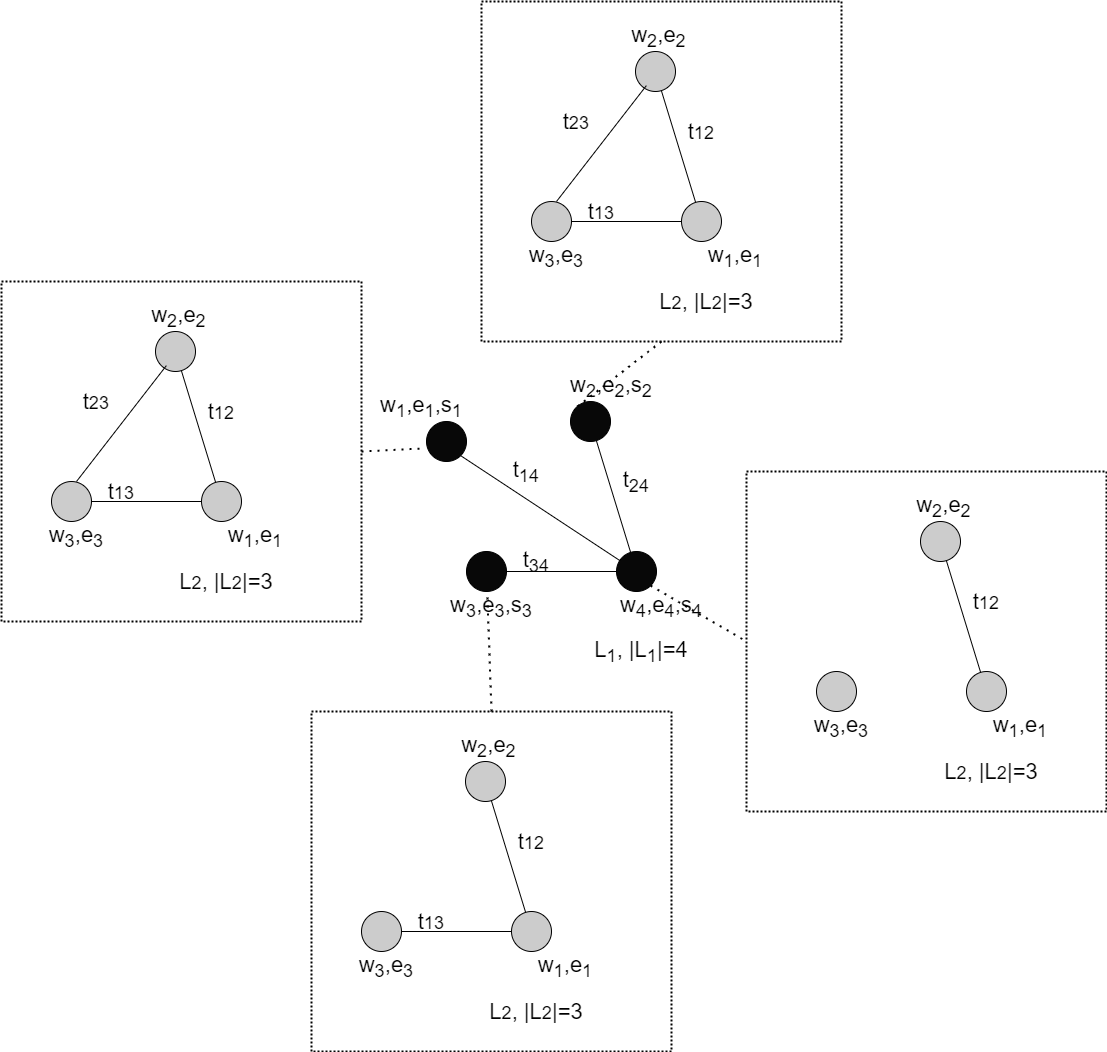}
\caption{Illustration of a DGNG modeled as graph, the first layer in this simple example consists of 4 units, the second layer (sub-DGNGs) of (maximal) 3 units. Units have parameters $(w,e,s)$, edges between units $u_i$ and $u_j$ are weighted with an age $t_{ij}$.}
\label{fig_schema_DGNG_layers}
\end{figure}

Formally, the structure of the DGNG is very similar to the RGNG model. It mirrors the two layered structure outlined in Section \ref{section_RGNG}, with the corresponding definitions of units, edges and functions. The explicit layer structure is illustrated in Figure \ref{fig_schema_DGNG_layers}. While building on a completely analoguous structure, the DGNG model conceptually follows a different interpretation and implementation compared to the RGNG:

On the first layer $L_1$ the input space is partioned in $N$ subregions, which correspond to certain prototypical input patterns learned by the neurons. Every partition is characterized by a single prototype vector at the center of the region. In a second step, every partition is now further divided via a sub-DGNG into $K$ subregions on the bottom layer $L_2$ , where $K=|L_2|$ is the number of modelled neurons. The units of the corresponding sub-DGNG hold vectors that represent the position relative to the center prototype. In terms of the neuron model perspective, every unit in the bottom layer effectivly describes the ``view'' of a neuron on this particular prototypical input pattern, the sub-DGNG therefore collects all variations of this pattern within the neuron group. This somehow turns the interpretation from the RGNG-based model upside down: The direct correspondence of neurons to units in the $L_1$ layer does not exist anymore, instead the representation of each neuron is distributed among the variations in the bottom layer $L_2$. In a certain sense, the information formerly contained in a direct neuron representation in the RGNG can be reconstructed by collecting all the perspectives of the selected neuron on the $N$ input partitions.

The main benefit of this approximation comes in terms of reduced computational complexity due to a largely reduced number of distance calculations per input. Instead of calculating the distance to every prototypical input pattern of every unit in $L_1$, which requires $O(n\cdot N \cdot K)$ operations, the calculation in the DGNG architecture only requires $O(n( N + K)$ operations \cite{Kerdels_Peters_2019} to obtain the BMUs: The algorithm simply has to check for the best partition ($N$ steps) and then calculates the distance for every neuron prototype within this partition ($K$ steps), eventually chosing the one with the smallest distance.

\vspace*{1ex} 

Experiments in \cite{Kerdels_Peters_2019} showed that this approximation indeed largely maintains the desired properties of the RGNG modeling, while being computationally more efficient.

\section{Methodology and dataset}\label{section_methodology} 

In this subsection the methodology for the capacity analysis is discussed, the corresponding model mechanisms and parameters are explained and the employed similarity metric is introduced. First notice that the concept of \textit{capacity} should not be confused with the \textit{classification} of the input images. For a maximum \textit{capacity} we expect the DGNG to assign a distinguishable activation signal to \textit{any} distinct input picture from the data set, no matter which class they belong to. However, at several points in the upcoming analysis a distinction between intra-class and inter-class similarity is made. This does not directly refer to the capacity itself but rather serves as a plausibility check, as one would expect, that images in the same class are usually ``more similar'' (i.e. cause a very similar activation signal) and therefore get confused more often.

The method conceptionally follows these steps: After training a DGNG on the training set, the algorithm is presented images from the testset (see Section \ref{section_Fashion_MNIST}) as an input. For each image, the ensemble activity from the network is calculated. This value depends on a contrast parameter $\gamma$, which is explained later in this section. The ensemble activities are then compared pairwise via a similarity metric (in this case, the \textit{Cosine Similarity} is used). If the similarity exceeds a certain threshold $\rho$, the images are considered to be identical by the network. The capacity is now determined by counting the number/fraction of pairwise distinguishable images from the testset.

In more formal detail, the procedure works as follows: The input image is presented to the DGNG as an input vector $\xi$, from which the algorithm calculates ensemble activities $\vec{a}=(a_0,\hdots ,a_{K-1})$ of the involved units. The ensemble activities are real vectors of the same size as $|L_2|=K$, mirroring an entry for every neuron perspective on the input. It is calculated from the $L_2$ prototypes of the two BMUs in the $L_1$ layer. The BMUs are determined by calculating the (Euclidean) distance between the image vector and the prototype of each $L_1$ unit. If the BMUs in the first layer are labeled $s_1$ and $s_2$, respectively, the activations $\hat{a}_i$ for every unit $i$ in $L_2$ are defined as:

\begin{equation}\label{eq_ensact1}
\hat{a}_i(\xi ,\gamma) := \gamma\left(1-\frac{||s_1.s.u_i.w - (\xi - s_1.w)||_2}{||s_2.s.u_i.w - (\xi - s_2.w)||_2} \right)\quad (i\in [0;K-1])\; .
\end{equation}

The notation $x.y$ denotes the element $y$ within the tuple $x$. From these activations the ensemble activities $a_i$ are obtained after a softmax operation:

\begin{equation}\label{eq_ensact2}
a_i := \frac{\exp{\left(\hat{a}_i\right)}}{\sum_{k=0}^{K-1}\exp{\left(\hat{a}_k\right)}}
\end{equation}

with $i\in [0;K-1]$.

This way, a $(28\times 28)$-pixel Fashion-MNIST input image is reduced to an activity vector of size $K$. In case the input vector $\xi$ is much closer to $s_1$ than to $s_2$, the activation in Eq. \ref{eq_ensact1} will come close to one and thus the ensemble activity Eq. \ref{eq_ensact2} will also approach one. In case the distance of $\xi$ to $s_1$ and $s_2$ is almost equal, the activation as well as the resulting ensemble activity will be close to zero. In other words, an ensemble activity close to one resembles a very specific activation of a DGNG prototype \cite{Kerdels_Peters_2018}. The parameter $\gamma$ can be interpreted as a contrast parameter. Increasing it amplifies high activations and thus shifts the similarities rather to the extreme limits \cite{Kerdels_Peters_2019}. As $\gamma$ has a large influence on the results, fixing techniques for the best parameter range are discussed in Section \ref{section_parameter_analysis}.

Following \cite{Kerdels_Peters_2019} the calculated ensemble activity vectors $(a_i, a_j)$ are compared by using the \textit{cosine similarity} $\text{cossim}(a_i, a_j)$. It is commonly defined as:

\begin{equation}\label{eq_cossim}
\text{cossim}(a_i, a_j) = \frac{a_i\cdot a_j}{|a_i|\cdot |a_j|} \in [0;1]\, .
\end{equation}

The maximum value of one indicates parallel input vectors, while zero indicates orthogonal vectors. Based on this metric the similarity threshold $\rho\in[0;1]$ is introduced. If the cosine similarity of the ensemble activities is below $\rho$, the image pair is considered non-identical by the DGNG. In this sense, a continuous similarity measure is transformed into a binary equality decision. The choice of $\rho$ therefore is of large importance for analysis: For a small value of $\rho$, more images are considered identical by the DGNG and sometimes it can not distinguish images, which are clearly distinguishable by standards of the human eye. Choosing a high value for $\rho$ will naturally result in the algorithm distinguishing more image pairs, but in extreme cases it might do so by focussing on minor details such as single pixels, which are not even recognizable by human standards. The choice of $\rho$ in context of balancing these tradeoffs is discussed and motivated in Section \ref{section_parameter_analysis}.

\section{The Fashion-MNIST dataset}\label{section_Fashion_MNIST}

Before conducting the analysis, the properties of the used dataset for training and evaluation are discussed in this section. The MNIST \cite{deng2012mnist} dataset has often been used as a testbed and illustration case for machine learning related tasks, i.e. classification. It consists of an overall of 70000 greyscale images in a format of 28x28 pixels, each depicting a handwritten number from zero to nine. The dataset is already divided in 60000 training images and 10000 testing images. It offers a simple opportunity to study structural differences between classes, as well as individual deviations of samples within the same class. Some preliminary analysis of DGNG properties have been conducted with MNIST in \cite{Kerdels_Peters_2019}. However, in this study MNIST is only used for calibration purposes and plausibility checks.

\begin{figure}[H]
\includegraphics[width=6cm]{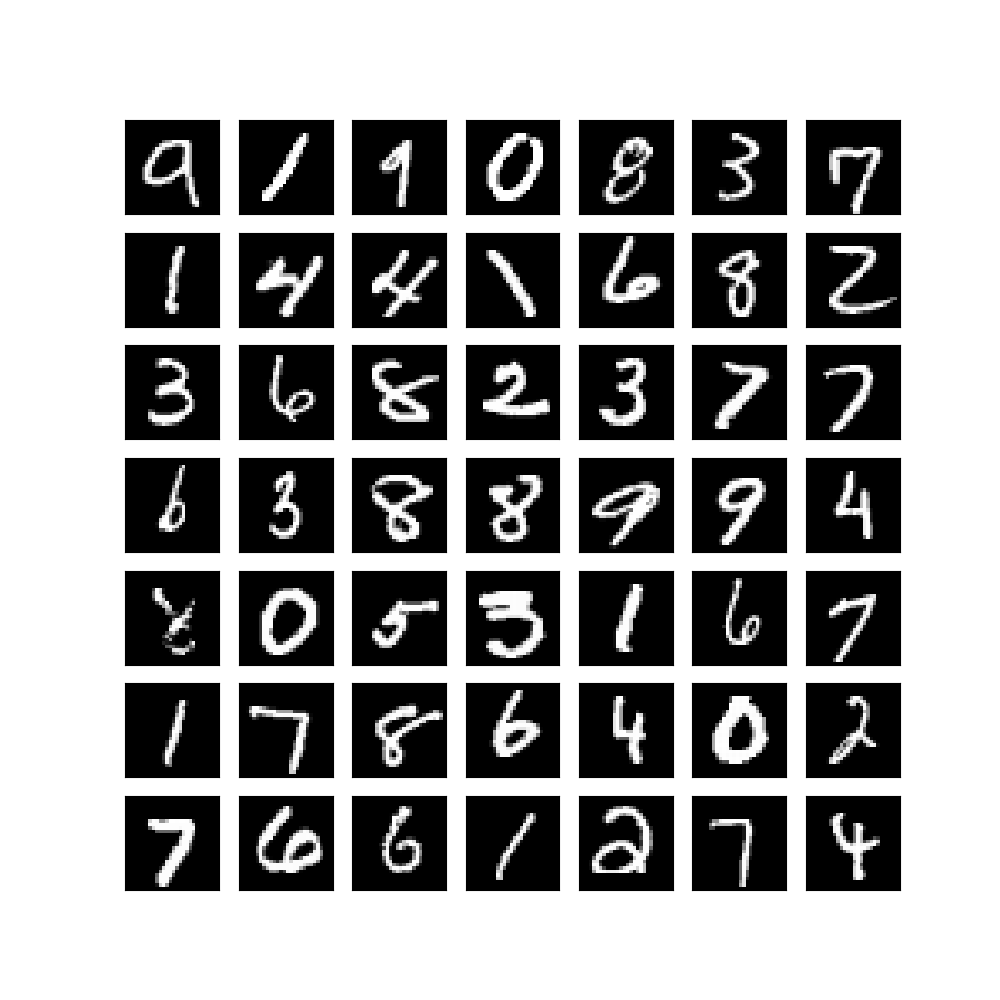}
\includegraphics[width=6cm]{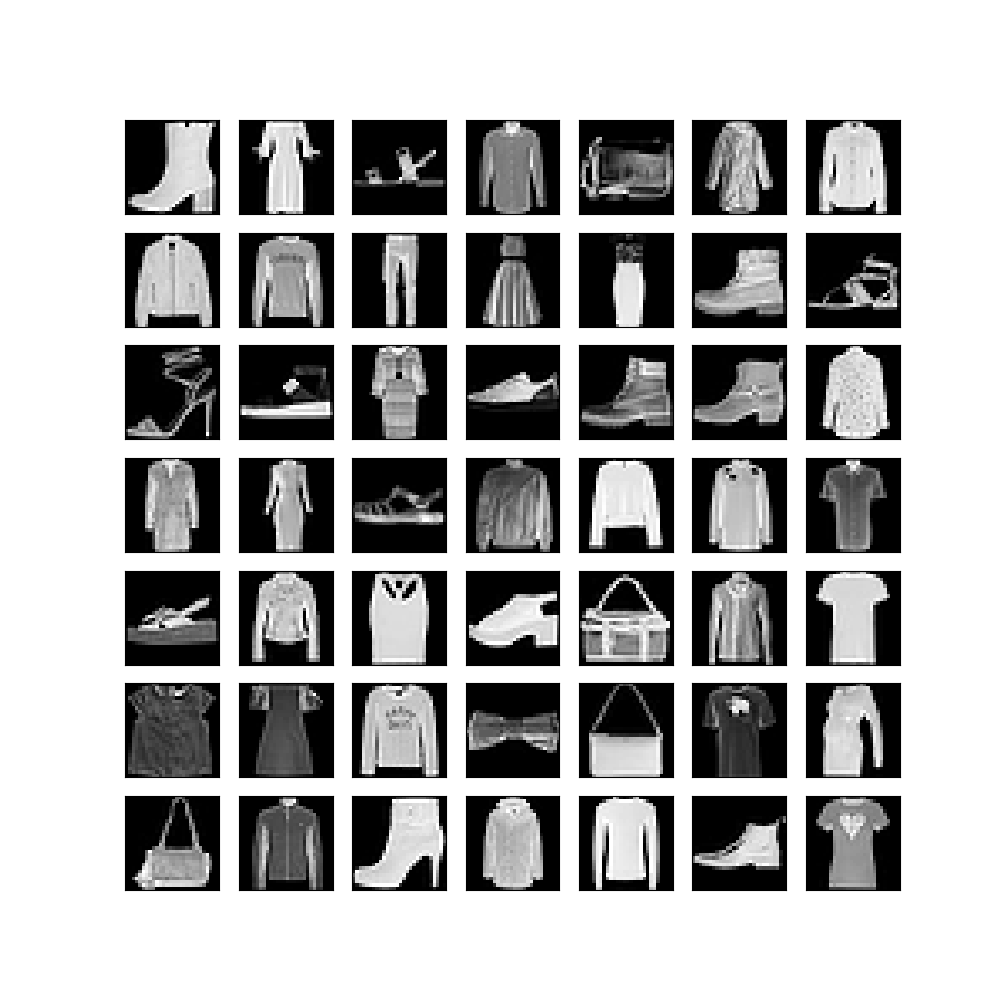}
\caption{\textit{Left:} Example images from the MNIST train dataset. \textit{Right:} Example images from the Fashion-MNIST train dataset.}
\label{fig_datasets}
\end{figure}

Instead, the more complex Fashion-MNIST  \cite{xiao2017fashionmnist} dataset is used for the capacity analysis. Fashion-MNIST contains fashion item pictograms, sorted into corresponding categories (shirt, shoe, etc.). On a structural level, the Fashion-MNIST mirrors the plain MNIST dataset completely: Both datasets have the same sample size (70000), number of categories (10) and contain samples in the format of 28x28 pixels. But on sample level, Fashion-MNIST offers an increased challenge for ML algorithms. The main reasons for this are the increased level of details, as well as the conceptional similarity between some of the categories, i.e. \textit{T-Shirt} (class 0) and \textit{Shirt} (class 6). An illustrative collection of examples from both datasets is given in Figure \ref{fig_datasets}. Using MNIST for calibration and Fashion-MNIST for extensive evaluation provides a decent opportunity to test the DGNG properties for more complex data, as well as probing the influence of data types on the evaluation performance.

\section{Parameter studies}\label{section_parameter_analysis}

In this section the parameters $\rho$ and $\gamma$ are examined to determine a proper evaluation range. 

\subsection{Determing the threshold parameter $\rho$}

As described in Section \ref{section_methodology}, the choice of the threshold parameter $\rho$ is somehow arbitrary. Still, an inappropriate choice can inhibit the network performance significantly. This relates to two adversive effects which have to be balanced: On the one hand, with a low threshold parameter the network cannot distinguishing clearly different images anymore, as it would accept even low similarity values as an identical input pair. On the other hand, chosing a too high treshold parameter might result in an overfitting effect, where the distinction process focusses on insignificant details (such as single pixel regions) rather than more plausible image features. In order to potentially model neurons in the context of cognitive functions, the network should be able to abstract features from data rather than just comparing single pixel values. An overview of arguments for parameter choice is depicted in Table \ref{tab_parameters}.

\begin{table}[H]
\centering
\begin{tabular}{||c||l|l||}
\hline\hline
&too low&too high\\ 
\hline\hline
\vspace*{0.5ex}
$\gamma$ &\pbox{5cm}{\vspace*{0.5ex}Activations in Equation \ref{eq_ensact1} are artificially shifted to zero, therefore ensemble activities are shifted to high values. Thus images become hard to distinguish based on ensemble activity. }&\pbox{5cm}{\vspace*{0.5ex}The resulting ensemble activities show a high contrast, they are either shifted to zero or to one. Therefore, the ability of DGNG to differentiate becomes increasingly coarse-grained. Low $\rho$ values therefore become less meaningful.}\\
\hline
\vspace*{0.5ex}
$\rho$&\pbox{5cm}{\vspace*{0.5ex}The number of distinguishable images is underestimated since already a similarity far below one is above the threshold and therefore considered too high to distinguish the images. In consequence, actually distinct images are considered not distinguishable.} &\pbox{5cm}{\vspace*{0.5ex}The number of distinguishable images is overestimated. Two images with a high similarity are still considered distinguishable, although the actual difference in the images might be small, even hardly recognizable by a human observer looking at the images. }\\
\hline\hline
\end{tabular}
\caption{Summary of the behavior of the number of distinguishable images for bounding cases of the empirically determined parameters $\gamma$ and $\rho$.}
\label{tab_parameters}
\end{table}

\begin{figure}[H]
\centering
\includegraphics[scale=0.3]{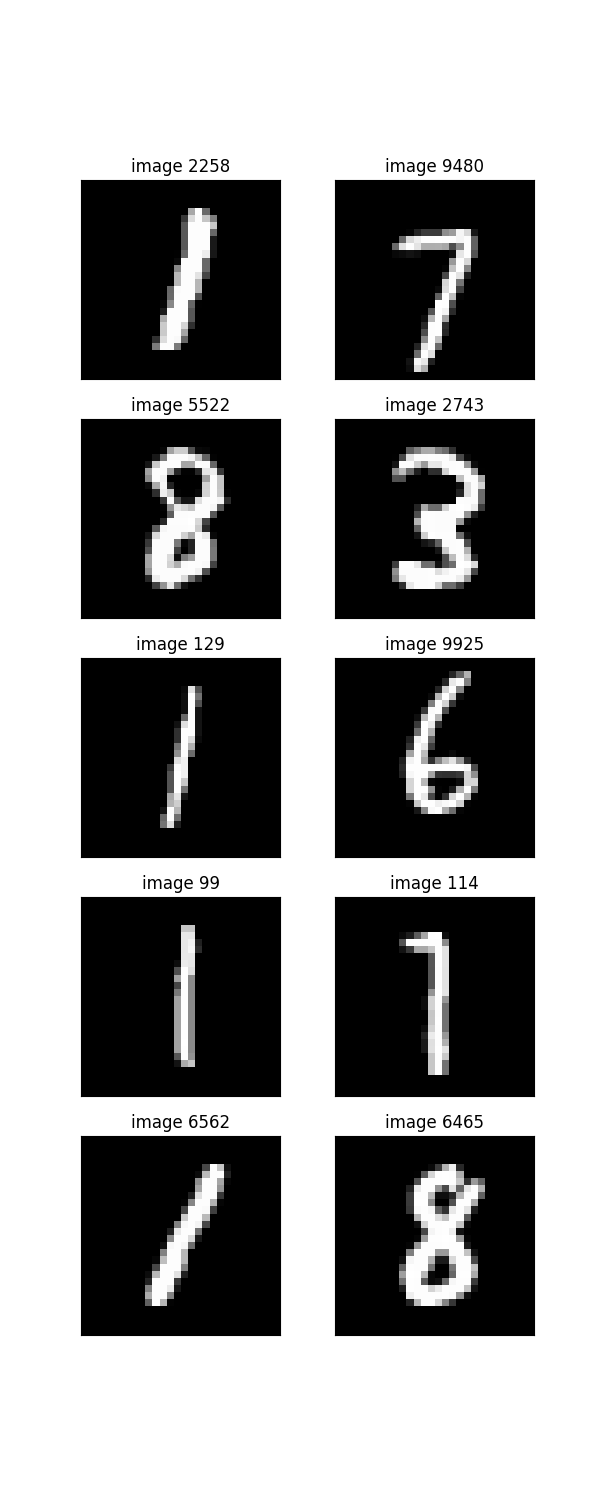}
\includegraphics[scale=0.3]{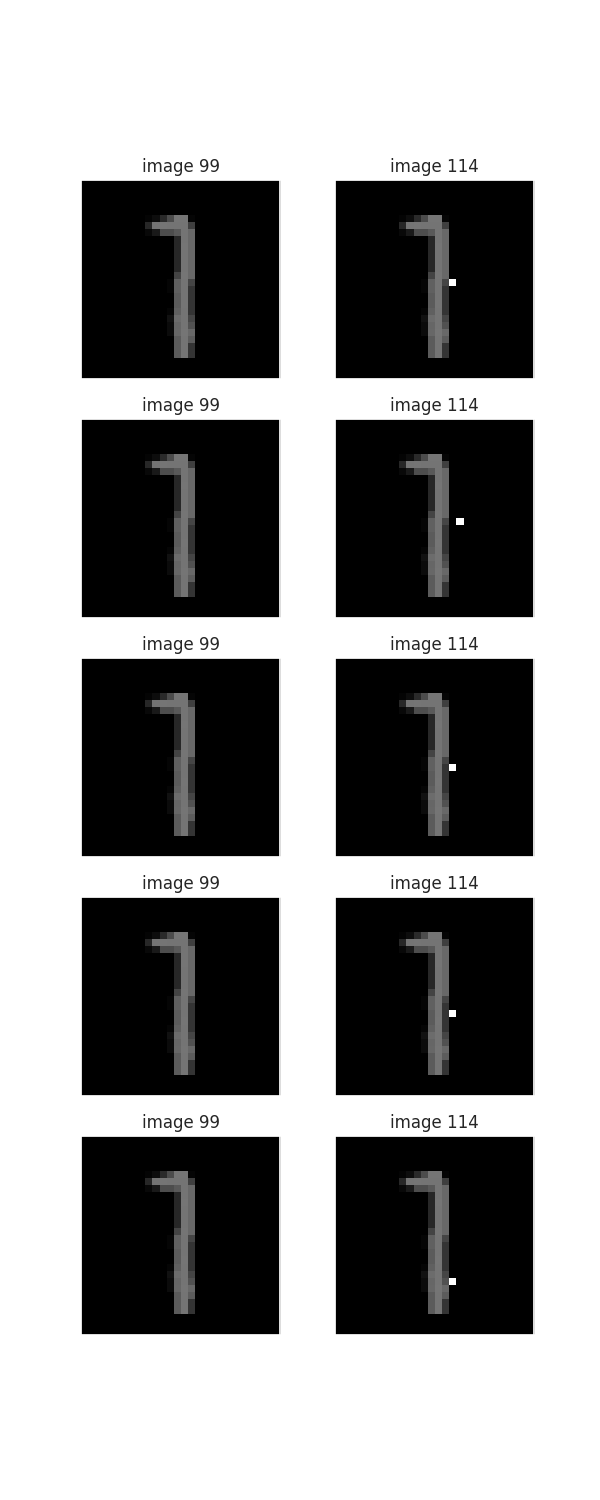}
\caption{\textit{Left:} Five exemplary MNIST image pairs (depicted on the left and right of the same row) from different classes which are not distinguished for an activation threshold $\rho=0.99$. \textit{Right:} Image 114 (original) and examples for single pixel manipulations of itself which lead to a decrease of similarity below 0.1. The grayscale of the images is shifted, the manipulated pixel is shown in white.}
\label{fig_mnist_confused}
\end{figure}

Based on these arguments, a pixel manipulation study is conducted to obtain a suitable range for $\rho$. If the network succesfully adopted plausible item features, the manipulation of a single pixel (or even a small pixel region) should not change the result of the distinction process. For the manipulation study a subset of five image pairs from the MNIST dataset is chosen. We select MNIST instances since it facilitates explaining the findings by looking at the images. Furthermore, since the dataset has much simpler structures, a single pixel manipulation is considered more severe and therefore having a stronger impact on the image similarity. In the base case analysis ($|L_1|=16$, $|L_2| = 100$) with $\gamma$ fixed from \cite{Kerdels_Peters_2019} these image pairs have a similarity value of over 0.99 while coming from different classes, thus being incorrectly considered identical. Exemplary image pairs are depicted in Figure \ref{fig_mnist_confused}.

The single pixel manipulation is conducted by changing every single pixel of an image to its extreme values of 0 (black) or 255 (white). In principle, a subset of 2 x 784 (image size) = 1568 manipulated images would emerge for every input image. In practical application however, images are omitted if the manipulation has no effect (i.e. if the pixel was already black or white). In most cases, the similarity of the pictures remains constant around 0.99, and in almost all cases the similarity remains above 0.95. Within our study scope, only for the images 2258 and 114 some single pixel manipulations lead to a larger similarity shift. In both cases, a ``7'' is compared to a ``1'', and at least for image 114, the distinction is not trivial even by human standards, see Figure \ref{fig_mnist_confused} (right). From this very prelimary analysis it is concluded, that both a threshold choice of 0.95 and 0.975 are suitable for ensuring a sufficient robustness against single pixel variations. For the following capacity studies all three values of 0.95, 0.975 and 0.99 are used.

Besides single pixel manipulation, also the variation of pixel areas (called a \textit{segment}) offers additional insights about the obtained feature representation capabilities (and therefore capacity) of the network. While an analysis in thorough detail is beyond the scope of this paper, a short plausability check is conducted. For this, the set of image pairs from the single pixel manipulation study is used. The images are divided into quadratic segments of 7 x 7 pixels. Analogue to the single pixel manipulation, the pixels in the selected segment are colored black as the background. For the MNIST dataset, this results in a subset of 16 images with non-overlapping segments for every input image. The corresponding similarity distributions are shown in Figures \ref{fig_mnist_segment_tg0_5522_2743} and \ref{fig_mnist_segment_tg1_5522_2743}. The results support the assumption that some regions of the image are characteristic for the similarity of their ensemble activities: The segment coverage can lead to a steep drop in the similarity, but not every coverage leads to a significant decrease. An illustrative example is shown in the images 5522 and 2743 (Figure \ref{fig_mnist_pixel_tg0_5522_2743}). The coverage of the bows for the ``3'' and ``8'' (or at the middle junction) lead to a decreasing similarity, while some coverages at the top of the bows or other regions have very little impact. 

\subsection{Determing the contrast parameter $\gamma$}\label{Section_gamma}

\begin{figure}[H]
\includegraphics[width=6cm]{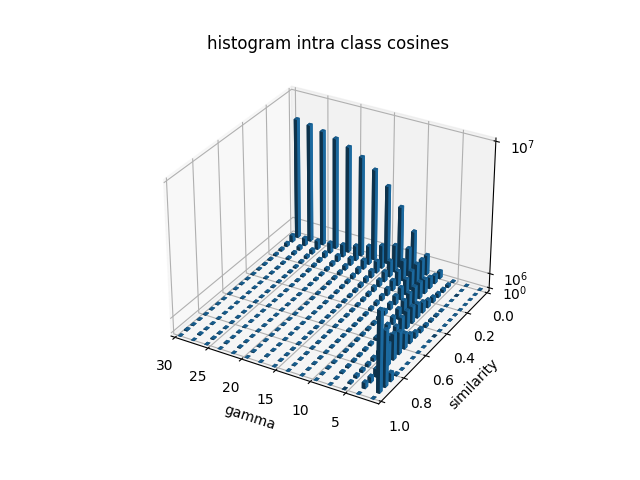}
\includegraphics[width=6cm]{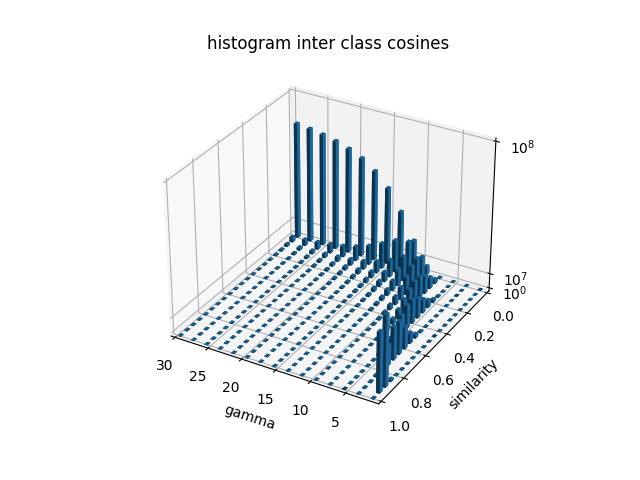}
\caption{Similarity histograms for each pair of images from the Fashion-MNIST test dataset, separately for image pairs from the same class \textit{(left)} or different classes \textit{(right)}. The similarity is binned, the bars show the count of each bin. The counting is done for a range of $\gamma$ values.}
\label{fig_Fashion-MNIST_histo}
\end{figure}

To obtain a suitable range for the contrast parameter $\gamma$ for the Fashion-MNIST dataset, the method from \cite{Kerdels_Peters_2019} is adopted. The procedure for capacity analysis outlined in Section \ref{section_methodology} is applied for $\gamma\in[1,31]$. The emerging similarity histograms are depicted in Figures \ref{fig_Fashion-MNIST_histo}. The key idea for determing $\gamma$ is to identify an interval in the histogram, where the similarity count is minimized. This interval refers to the best capabilities for distinguishing images for this parameter value. In the case at hand, the minima for intra- and interclass similarities are obtained at $\gamma_{intra}\in[11;15]$ and $\gamma_{inter}\in[7;11]$, respectively. Throughout the study, $\gamma$ is individually fixed via this procedure for every layout.

\section{Capacity analysis for Fashion-MNIST dataset}\label{section_analysis}

In this section, the capacity analysis is conducted by using the method described in Section \ref{section_methodology} as well as the parameter selection outlined in Section \ref{section_parameter_analysis}. As a starting point, the base case configuration is examined. From there on, the effects of varying the layer sizes are studied. All evaluations have been conducted at 10 million training cycles if not stated otherwise. 

\subsection{Base case analysis and illustration}

\begin{figure}[H]
\includegraphics[scale=0.5]{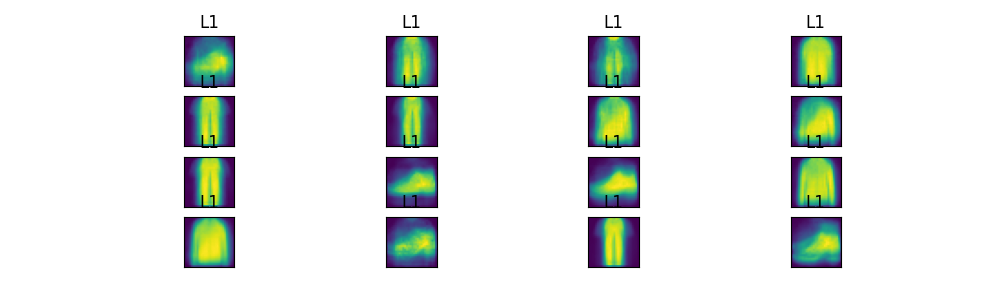}
\includegraphics[scale=0.5]{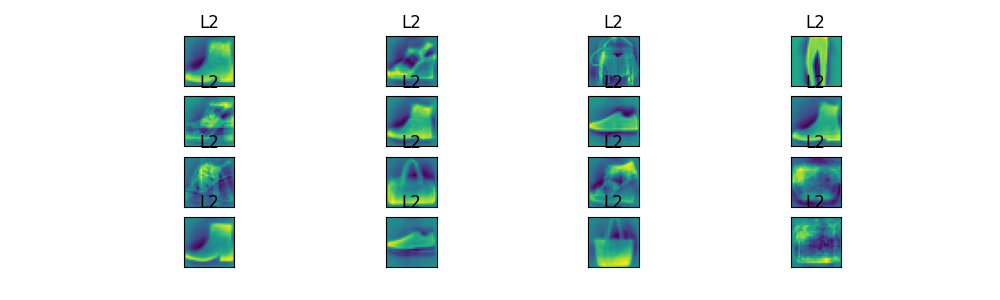}
\caption{\textit{Upper row:} Image plot of the 16 prototypes learned by the $L_1$ layer of the DGNG. \textit{Lower row:} Image plot of 16 prototypes learned in the $L_2$ layer, more specifically by the neurons in the sub-DGNG of the first prototype in $L_1$. Both are from the DGNG with reference architecture trained on the Fashion-MNIST dataset.}
\label{fig_proto}
\end{figure}

The base case analysis refers to the same network architecture used in \cite{Kerdels_Peters_2019} i.e. $|L_1|=16$, $|L_2|=100$. Figure \ref{fig_proto} shows the visualization of the learned prototypes in the $L_1$ and $L_2$ layers, where the exemplary visualization for the $L_2$ layer refers to the first prototype of the $L_1$ layer. The visualization shows a plausible agreement with the expectation, the prototypes optically correspond to recognizable features from the image categories. Note, that the learned prototypes sometimes are combinations of fashion items, for example the second item in the first row of Figure \ref{fig_proto} shows a combination of ``shirt'' and ``shoe''. Thus this prototyp should show an high activation for pictograms of both categories. The $L_2$ prototypes show a much sharper defined picture, illustrating the principle network hierarchy: While the $L_1$ prototypes reflect a rough partition (and therefore some kind of classification), the $L_2$ prototypes quantify small variations of the linked $L_1$ prototype.

In terms of capacity, results of 39\% ($\rho=0.95$) and 65\% ($\rho=0.975$) are achieved (see Table \ref{app_FMNIST_results_L2var}). It is again emphasized that this refers to the amount of images, which are not seen as identical with \text{any} other image from the test set.

\subsection{Capacity analysis of various layer sizes}

To probe the scaling properties of the network, the evaluation is repeated for different network sizes. In principle, both network layers $L_1$ and $L_2$ can be varied and in both cases, a positive impact on the capacity is expected for increasing layer size. To analyze these effects in detail, a stepwise increase of both layer sizes is conducted. The analysis is conducted for $\rho\in[0.95, 0.975, 0.99]$ and the contrast parameter $\gamma$ is determined individually for every configuration according to the procedure outlined in Section \ref{Section_gamma}. The result of an evaluation fluctuates within a certain range due to initial conditions and the number of training cycles. These fluctuations are discussed in Section \ref{section_variance}, while in the following illustrations the average of these results is depicted.

\subsection{Variation of $|L_2|$}\label{section_L2_variation}

\begin{figure}[H]
\includegraphics[scale=0.4]{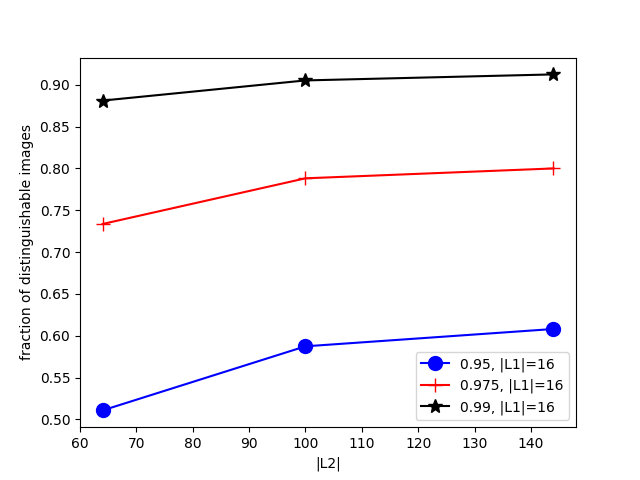}
\includegraphics[scale=0.4]{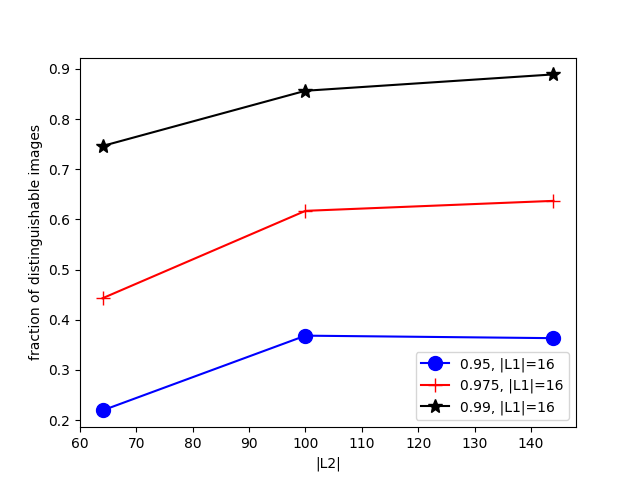}
\caption{Plot of the fraction of the maximum  distinguishable images for $|L_1|=16$, $\rho = 0.95$, $0.975$, and $0.99$ with individually fixed $\gamma$ over the examined layer sizes $|L_2|\in[64,144]$. The lines are drawn only as guide to the eyes. \textit{Left:} MNIST, \textit{Right:} Fashion-MNIST averaged over varying training cylce number}
\label{figure_L2_variation_L1_16_avg}
\end{figure}

In a first step, the size of the second layer is varied in the interval of $|L_2| \in [64,100,144]$ while for the first layer the base case size $|L_1|=16$ is kept. The results for some evaluation runs in terms of the fraction of the maximum  distinguishable images are visualized in Figure \ref{figure_L2_variation_L1_16_avg}, along with the MNIST results for comparison. The full results in detail can be found in Table \ref{app_FMNIST_results_L2var}.

While the performance initially shows a significant improvement by increasing the size of the $L_2$ layer, the effect flattens out above $|L_2|=100$ for both MNIST and Fashion-MNIST. This hints that some kind of saturation has occured: All substantial features of the data have already been described sufficiently by the network at this point and adding new units to the $L_2$ layer does therefore not significantly improve the performance. 

\subsection{Variation of $|L_1|$}\label{section_L1_variation}

\begin{figure}[H]
\includegraphics[scale=0.4]{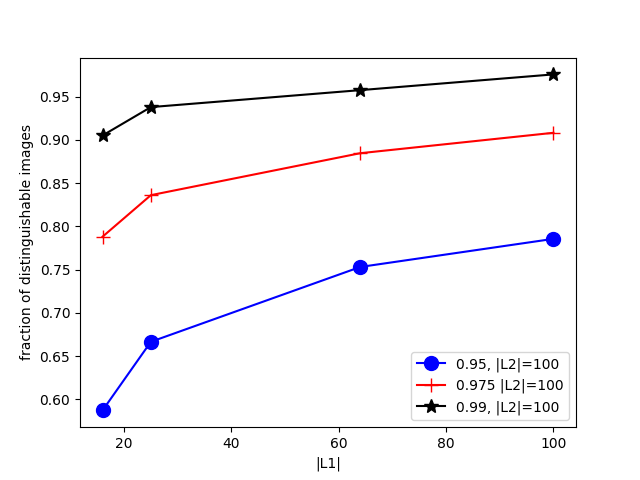}
\includegraphics[scale=0.4]{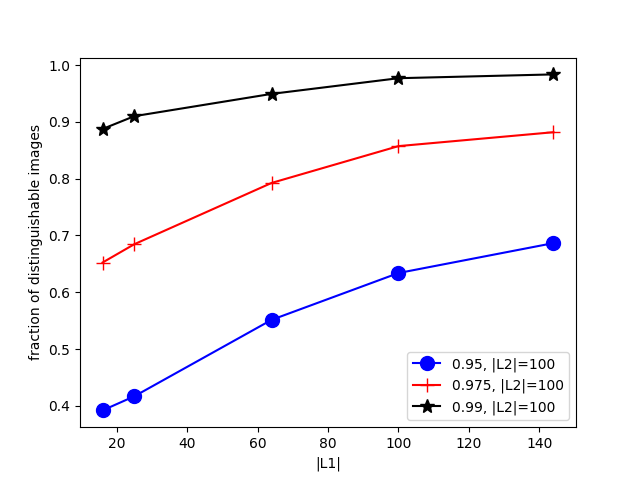}
\caption{Plot of the fraction of the maximum  distinguishable images for $\rho = 0.95$, $0.975$, and $0.99$ over the examined layer sizes $|L_1|$. The lines are drawn only as guide to the eyes. \textit{Left:} MNIST, \textit{Right:} Fashion-MNIST}
\label{fig_L1_variation_L2_100}
\end{figure}

In a second analysis step the number of units in the first layer $L_1$ is increased stepwise. In total, five configurations are tested: $|L_1| = 16, 25, 64, 100$ and $144$. The size of the second layer is fixed at the base case value $|L_2|=100$ for all analysis steps. Figure \ref{fig_L1_variation_L2_100} depicts the fraction of distinguished image pairs from the test set for the different network architectures. The detailed numbers for every analysis step can be found in Table \ref{app_FMNIST_results_L1var}.\\

As expected, the fraction of distinguished image pairs increases by adding more units to the $L_1$ layer. In the region $|L_1|>100$, the system also shows signs of saturation as the capacity performance benefit for newly added units decreases. This saturation is probably connected to the level of detail of the corresponding data set: Once the categorizational variety of the set items is nearly fully captured, introducing additional categories does not aid the construction of useful features anymore. This is emphasized by a comparison to the MNIST case, which is also depicted in Figure \ref{fig_L1_variation_L2_100}. The saturation also occurs in this case, but due to the decreased complexity compared to Fashion-MNIST, it already sets in between $|L_1|=25$ and $|L_1|=64$, meaning that the categorizational variety has been covered with a smaller size of $L_1$.

\subsection{Convergence properties and variance of results}\label{section_variance}

\begin{figure}[H]
\includegraphics[scale=0.4]{bilder/verlaufsplot_mnist_L1_16__L2_variation.png}
\includegraphics[scale=0.4]{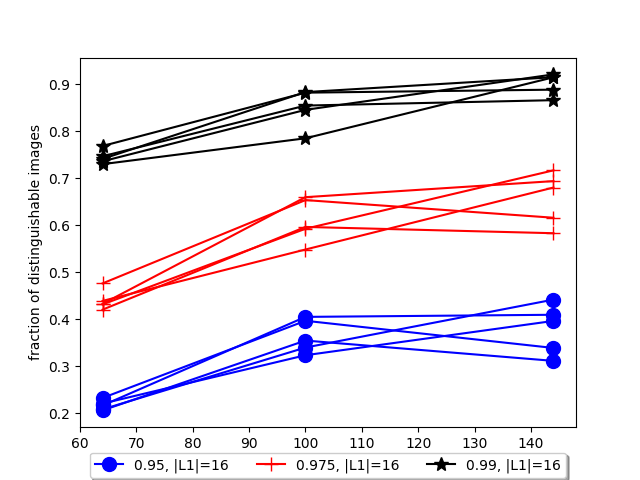}
\caption{Illustration of capacity oscillation for $|L_1|=16$, $\rho = 0.95$, $0.975$, and $0.99$ with individually fixed $\gamma$ over the intervall $|L_2|\in[64,144]$. The lines are drawn only as guide to the eyes. \textit{Left:} MNIST, \textit{Right:} Fashion-MNIST}
\label{figure_L2_variation_L1_16}
\end{figure}

As mentioned in Section \ref{section_L2_variation}, the results for the capacity might vary according to training cycles and initial conditions of the evaluation (e.g. seed of the random number generator). An illustration for this is shown in Figure \ref{figure_L2_variation_L1_16}. A more quantitative picture of the fluctuations is given in the Tables \ref{app_seed_variation} and \ref{app_cycle_variation} in Appendix \ref{app_variation}, where the results according to  random seed variation and the number of training cycles have been collected. In both cases, the fluctuation amounts to about 300-450 distinguished images (corresponding to 3-4.5\% of the testset size).

The reason for this is the convergence behavior of the DGNG. For the purpose of multi-run evaluation, the learning rate for the network has been kept rather large (at around 1000). Once the prototypes in the network have mostly converged, it will oscillate between some configurations close to a hypothetical optimum. Note, that the variation due to training cycle and initial conditions are both two sides of the same effect: The ''oscillation state`` chosen for evaluation depends on the end point of the training process.

In practical applications this oscillation behavior can be avoided by either choosing a smaller learning rate (increasing the training time correspondingly) or decreasing the learning rate over time in later training cycles (although the latter method will not always converge to the same configuration). For the purpose of this study, the oscillating behavior was maintained and described to give an impression of the variation range and statistical uncertainty of the results. As can be seen in Figure \ref{figure_L2_variation_L1_16} the variation width does not contradict the qualitative assessment of the result.

\section{Conclusion and Outlook}\label{section_conclusion}

The purpose of this paper was to study the capacity related properties of a DGNG architecture on the Fashion-MNIST dataset. After the methodology and the corresponding metrics have been established in Section \ref{section_methodology}, the properties of the dataset have been discussed in Section \ref{section_Fashion_MNIST}. Following a parameter study and plausability check with pixel variation studies (see Section \ref{section_parameter_analysis}), in Section \ref{section_analysis} the capacity studies have been conducted for varying layer sizes of the DGNG.

The maximum capacity of ca. 88\% (See Table \ref{app_FMNIST_results_L1var}) has been found for the configuration $|L_1|=144, |L_2|=100$, which as expected is also the configuration with the largest number of units examined in this study. Note that there are even larger capacity values for $\rho =0.99$, but this cases have been ruled out for reasons explained in Section \ref{section_parameter_analysis}.
The analysis confirms the expectation that the capacity performance depends on the complexity of the input data. If for example the second layer is fixed at the base case $|L_2| = 100$ and the threshold parameter is selected to $\rho=0.975$, for the MNIST dataset 25 units in the $L_1$ layer are sufficient to reach about 84\% (see Table \ref{app_MNIST_results_L1var}) capacity. For the Fashion-MNIST dataset however, a comparable performance of about 86\% (see Table \ref{app_FMNIST_results_L1var}) is only achieved for 100 units in the $L_1$ layer. The same observation holds for the $|L_2|$ variation as well as other treshold parameters.

The variation study of the $L_1$ layer (see Figure \ref{fig_L1_variation_L2_100}) shows a saturation effect at larger regions of $|L_1|$ for both data sets, MNIST and Fashion-MNIST. This saturation is interpreted as a consequence of the function of the $L_1$ layer within the architecture: Once enough prototypes are available to parametrize the major differences between the different image categories, additional units do not contribute to form meaningful prototypes anymore. In terms of the neural gas model, this means once the prototypes are distributed among the whole input space in a certain density, additional units do not cover significant portions of empty space. This results hint that a comparativly small neuron group in this model is able to capture large portions of information, while inflating the  numbers of units in the first layer (corresponding to their dendritic tree) beyond a certain point will have gradually less effect.

The variation of the $L_2$ layer revealed a similar effect (see Figure \ref{figure_L2_variation_L1_16_avg}): As expected, the performance in principle benefits from adding new units to the layer, but saturation sets in after around $|L_2|=100$. Again similar to the $L_1$ variation this corresponds to a situation in which the $L_2$ units sufficiently covered most of the variation associated to a certain input partition. It is expected, that the saturation point correlates with the complexity of the data set.

Comparing both variations it is concluded, that the steepest rise in capacity performance is achieved by increasing $|L_1|$ (if the saturation point is not yet reached). The layer size of $L_2$ on the other hand has a much smaller influence on the performance. This can formally be explained by the layer architecture itself as increasing the $L_1$ layer adds more neurons to the network than increasing the $L_2$ layer (if, as usually the case, $|L_2| > |L_1|$).

In summary, the capacity study showed that DGNG architectures are able to uniquely store data patterns by abstracting features of the data sets in terms of prototypes. The scaling properties show that for datasets with some complexity (as the Fashion-MNIST) modest capacity values of about ~65\% can already be achieved with small setups ($|L_1|=16, |L_2|=100$). Future work might include a more detailed study of pixel/segment variation to obtain a better understanding of the abstraction capabilities of the architecture. Also, probing even more complex datasets with larger architectures can reveal more information about scaling properties, computational efficiency and prospects of application in scientifical modeling.

\begin{bibliography}{lit}
\bibliographystyle{unsrt}
\end{bibliography}

\appendix

\section{Detailed results tables and plots}

In this Section some tables for more detailed numerical results of the evaluation are collected. All evaluations are done with 10 million training cycles if not stated otherwise.

\subsection{Evaluation of varying $L_1$ size}

\begin{table}[H]
\centering
\begin{tabular}{|c||c|c|c|}
\hline
$|L_1|,|L_2|=100$&$\rho =0.95$&$\rho=0.975$&$\rho=0.99$\\
\hline\hline
16&5873&7881&9051\\
25&6666&8361&9379\\
64&7531&8846&9575\\
100&7856&9082&9757\\
\hline
\end{tabular}
\caption{Amount of uniquely distinguished images from the MNIST testset of 10000 images for various $L_1$ layer sizes and selected values of $\rho$. The second layer is fixed at the base case $|L_2|=100$.}
\label{app_MNIST_results_L1var}
\end{table}

\begin{table}[H]
\centering
\begin{tabular}{|c||c|c|c|}
\hline
$|L_1|,|L_2|=100$&$\rho =0.95$&$\rho=0.975$&$\rho=0.99$\\
\hline\hline
16&3919&6523&8875\\
25&4167&6845&9100\\
64&5514&7925&9493\\
100&6338&8573&9769\\
144&6865&8818&9836\\
\hline
\end{tabular}
\caption{Amount of uniquely distinguished images from the Fashion-MNIST testset of 10000 images for various $L_1$ layer sizes and selected values of $\rho$. The second layer is fixed at the base case $|L_2|=100$.}
\label{app_FMNIST_results_L1var}
\end{table}

\subsection{Evaluation of varying $L_2$ size}

\begin{table}[H]
\centering
\begin{tabular}{|c||c|c|c|}
\hline
$|L_2|$, $|L_1|=16$&$\rho =0.95$&$\rho=0.975$&$\rho=0.99$\\
\hline\hline
64&5107&7335&8810\\
100&5873&7881&9051\\
144&6080&7999&9123\\
\hline
\end{tabular}
\caption{Amount of uniquely distinguished images from the MNIST testset of 10000 images for various $L_2$ layer sizes and selected values of $\rho$. The first layer is fixed at the base case $|L_1|=16$.}
\label{app_MNIST_results_L2var}
\end{table}

\begin{table}[H]
\centering
\begin{tabular}{|c||c|c|c|}
\hline
$|L_2|$, $|L_1|=16$&$\rho =0.95$&$\rho=0.975$&$\rho=0.99$\\
\hline\hline
64&2346&4589&7555\\
100&3919&6523&8875\\
144&2745&5292&8285\\
\hline
\end{tabular}
\caption{Amount of uniquely distinguished images from the Fasion-MNIST testset of 10000 images for various $L_2$ layer sizes and selected values of $\rho$. The first layer is fixed at the base case $|L_1|=16$.}
\label{app_FMNIST_results_L2var}
\end{table}

\subsection{Pixel/segment manipulation plots}

In this Section exemplary distribution plots for the single pixel and segment manipulation studies are collected.

\begin{figure}[H]
\hspace{-5ex}
\includegraphics[width=7cm]{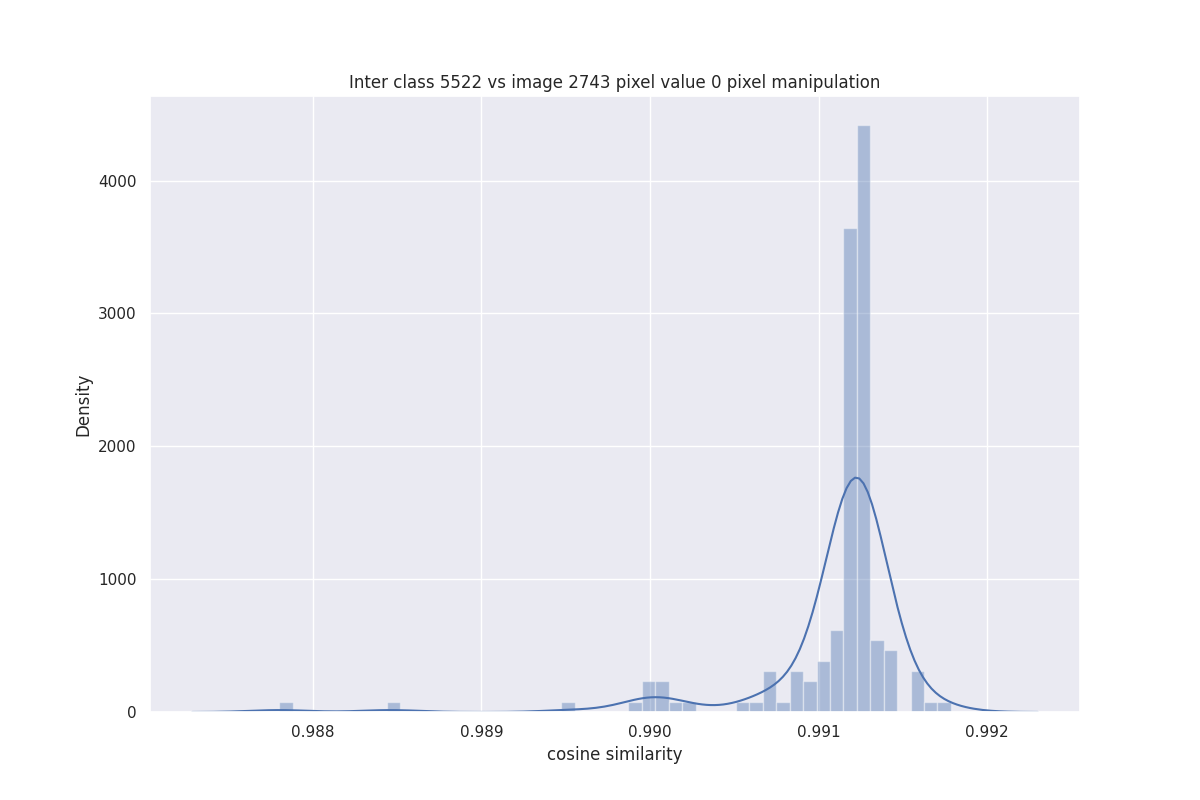}
\hspace{-2ex}
\includegraphics[width=7cm]{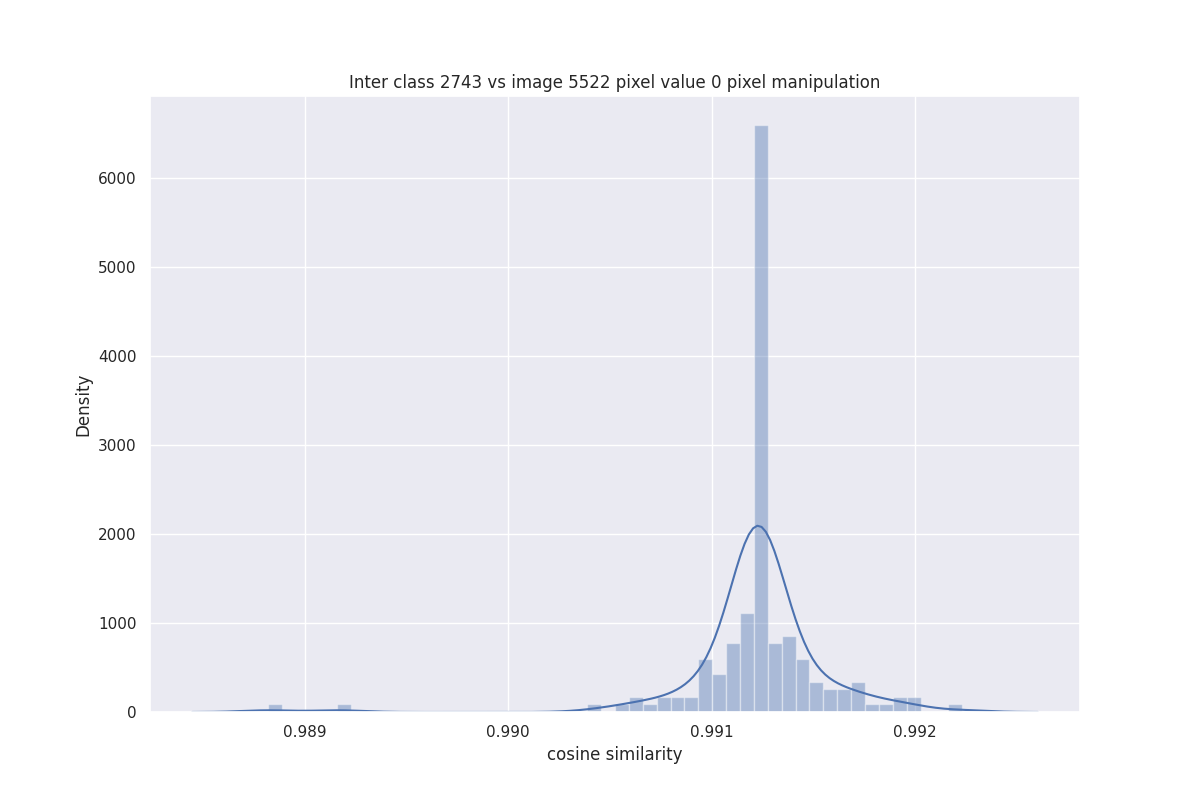}
\caption{Distribution of the similarity between image pairs 5522 and 2743 with single pixel manipulations to zero in image 5522 \textit{(left)} and image 2743 \textit{(right)}, respectively.}
\label{fig_mnist_pixel_tg0_5522_2743}
\end{figure}

\begin{figure}[H]
\hspace{-5ex}
\includegraphics[width=7cm]{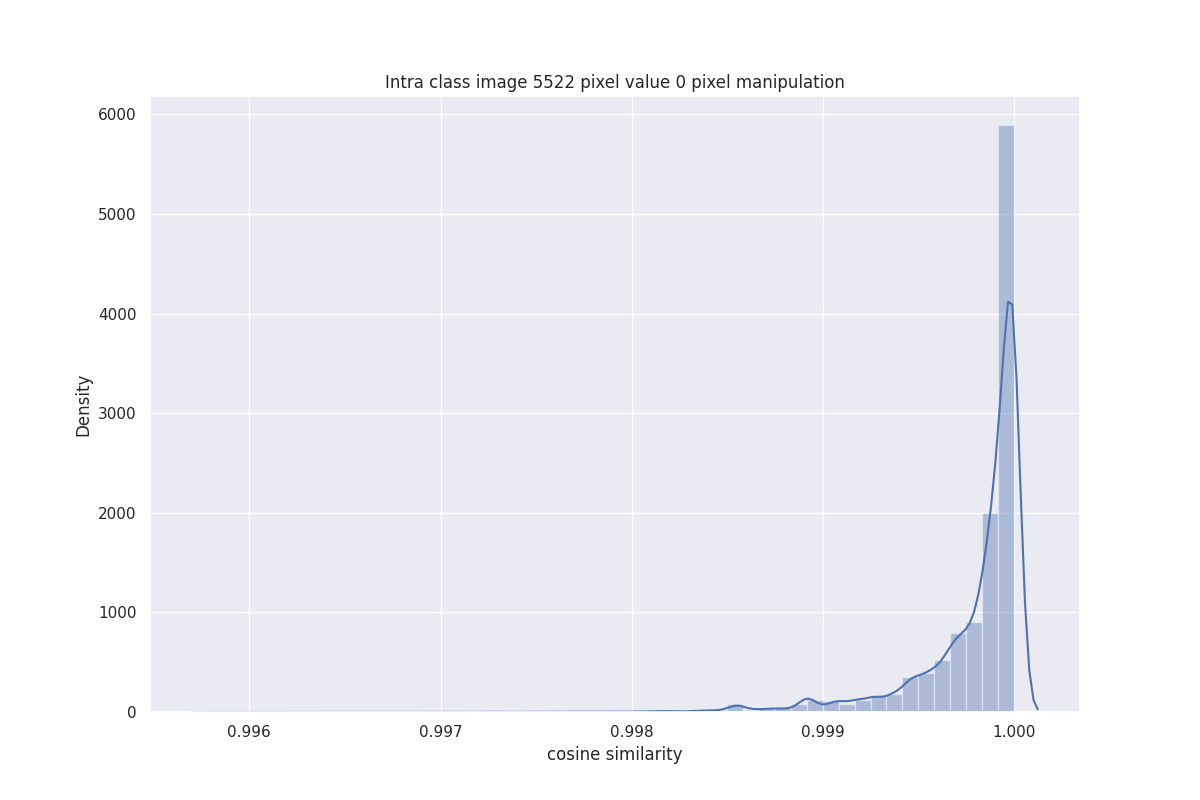}
\hspace{-2ex}
\includegraphics[width=7cm]{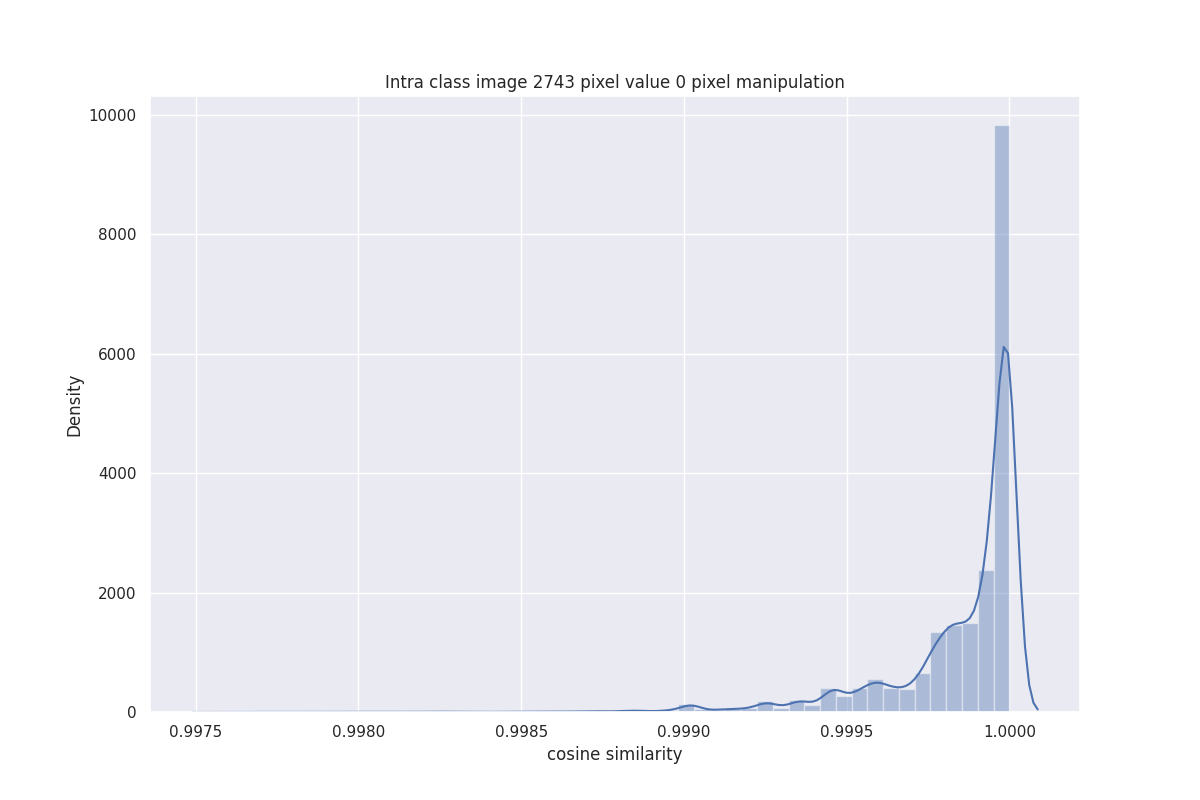}
\caption{Distribution of the similarity of original image 5522 \textit{(left)} and 2743 \textit{(right)}, respectively, with itself after single pixel manipulations to zero.}
\label{fig_mnist_pixel_tg1_5522_2743}
\end{figure}

\begin{figure}[H]
\hspace{-5ex}
\includegraphics[width=7cm]{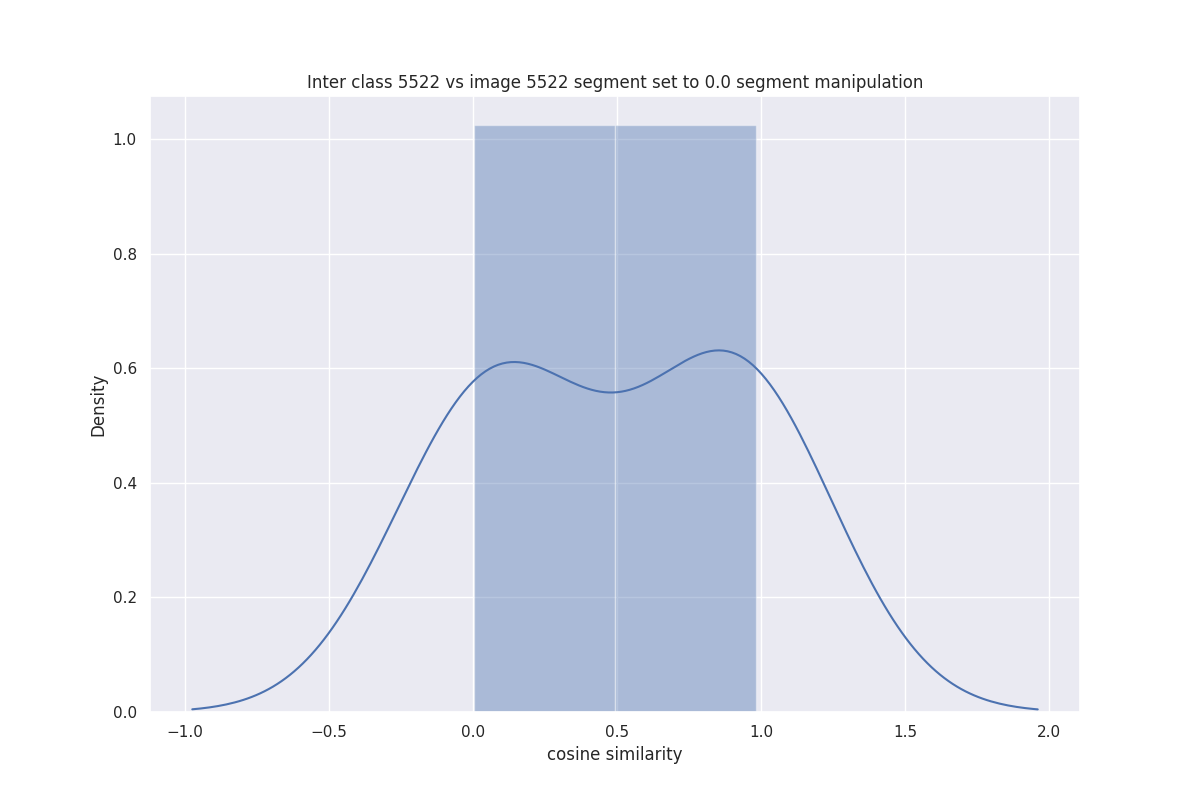}
\hspace{-2ex}
\includegraphics[width=7cm]{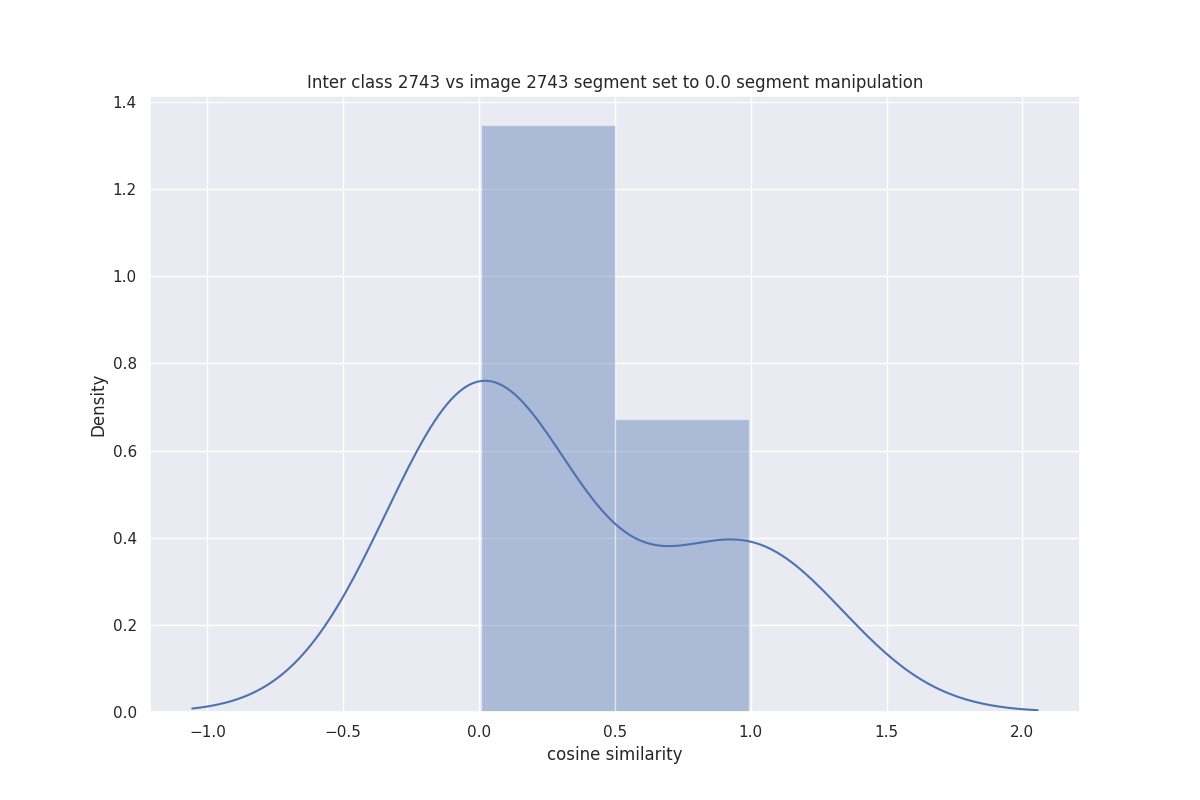}
\caption{Distribution of the similarity between image pairs 5522 and 2743 with segment manipulations in 5522 to zero \textit{(left)} and the same for image 2743 \textit{(right)}, respectively.}
\label{fig_mnist_segment_tg0_5522_2743}
\end{figure}

\begin{figure}[H]
\hspace{-5ex}
\includegraphics[width=7cm]{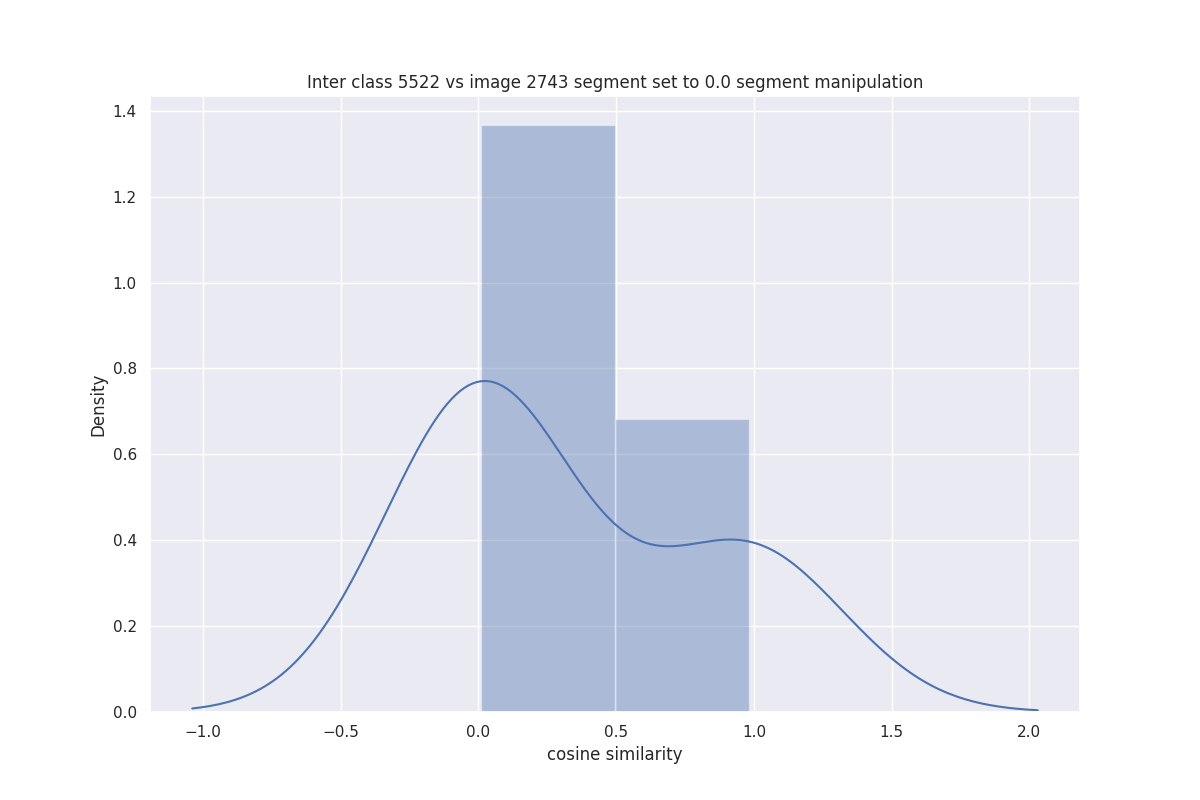}
\hspace{-2ex}
\includegraphics[width=7cm]{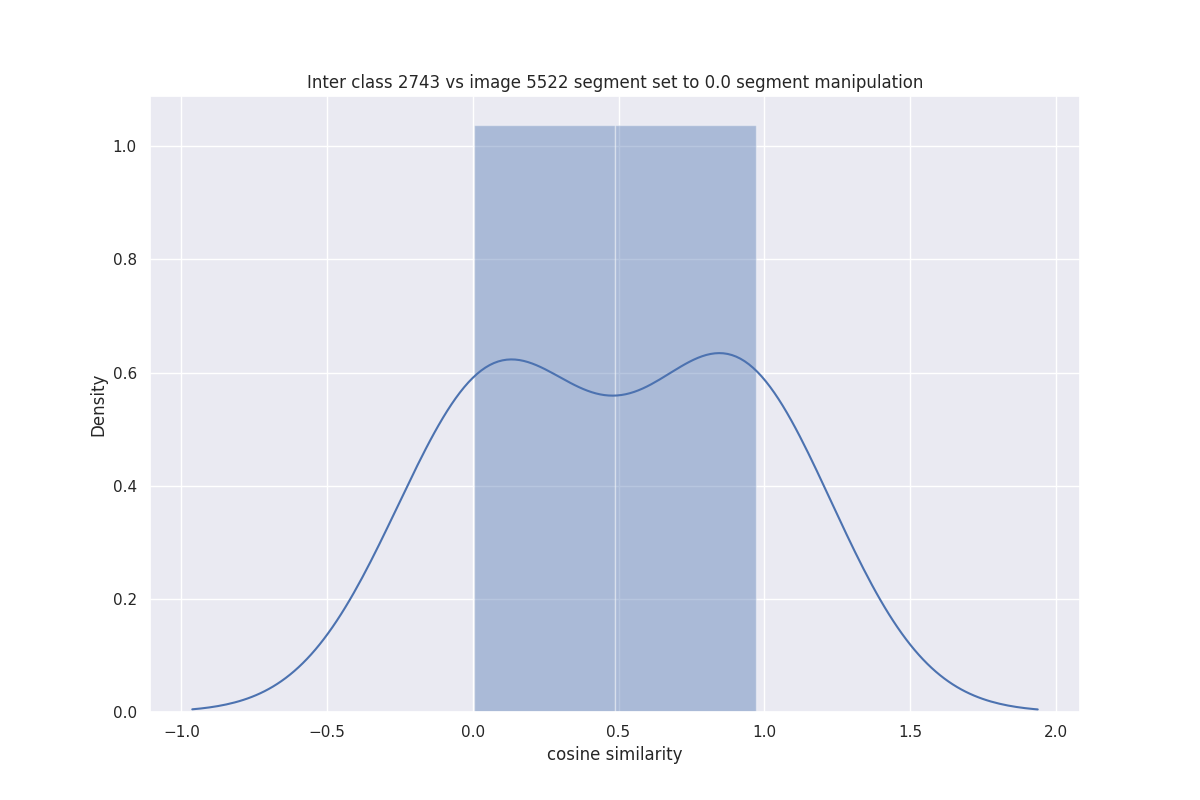}
\caption{Distribution of the similarity of original image 5522 with 2743 with segments were covered \textit{(left)} and vice versa \textit{(right)}.}
\label{fig_mnist_segment_tg1_5522_2743}
\end{figure}

\subsection{Analysis of fluctuations}\label{app_variation}

Two sources of statistical variation are investigated: Varying training cycle size (between 10 million and 15 million in steps of 1 million) and varying random seed for fixed training cycle length (15 million cycles). The average value is denoted by $\mu$ and the empirical variance by $\sigma^2$.

\begin{table}[H]
\centering
\begin{tabular}{|c||c|c|c|}
\hline
$|L_2|=100$, $|L_1|=16$&&&\\
\hline\hline
Seed&$\rho =0.95$&$\rho=0.975$&$\rho=0.99$\\
\hline\hline
base case&	3231&	5483&	7852\\
142&	3300&	5882&	8559\\
742&	3399&	5993&	8600\\
442&	2849&	5281&	8342\\
542&	2938&	5332&	8151\\
\hline
$\mu$&3143&5594&8301\\
$\sigma$ & 238&324&309\\
\hline
\end{tabular}
\caption{Evaluation results for the Fashion-MNIST dataset on base case architecture and selected threshold parameters depending on various random seeds.}
\label{app_seed_variation}
\end{table}

\begin{table}[H]
\centering
\begin{tabular}{|c||c|c|c|}
\hline
$|L_2|=100$, $|L_1|=16$&&&\\
\hline\hline
Cycles (million)&$\rho =0.95$&$\rho=0.975$&$\rho=0.99$\\
\hline\hline
10&	3919&	6523&	8875\\
11&	3544&	5964&	8549\\
12&	3965&	6536&	8824\\
13&	4048&	6598&	8836\\
14&	3395&	5925&	8458\\
15&	3231&	5483&	7852\\
\hline
$\mu$&3684&6172&8566\\
$\sigma$&339&451&389\\
\hline
\end{tabular}
\caption{Evaluation results for the Fashion-MNIST dataset on base case architecture and selected threshold parameters depending on training cycles.}
\label{app_cycle_variation}
\end{table}

\end{document}